\pdfoutput=1

\documentclass[11pt]{article}

\usepackage[final]{acl}

\usepackage{times}
\usepackage{latexsym}
\usepackage{multirow}
\usepackage{pifont}
\usepackage{marvosym}

\usepackage[T1]{fontenc}

\usepackage[utf8]{inputenc}

\usepackage{microtype}

\usepackage{inconsolata}

\usepackage{graphicx}

\title{NavRAG: Generating User Demand Instructions for Embodied Navigation through Retrieval-Augmented LLM}

\author{
  \textbf{Zihan Wang\textsuperscript{2}},
  \textbf{Yaohui Zhu\textsuperscript{3}},
  \textbf{Gim Hee Lee\textsuperscript{2}},
  \textbf{Yachun Fan\textsuperscript{1\ \Letter}}
\\
  \textsuperscript{1}School of Artificial Intelligence, Beijing Normal University \\
  \textsuperscript{2}School of Computing, National University of Singapore \\
  \textsuperscript{3}College of Information Science and Technology, Beijing University of Chemical Technology
\\
  \small{ 
  zihan.wang@u.nus.edu,
  fanyachun@bnu.edu.cn
  }
}

\begin{document}
\maketitle
\begin{abstract}

Vision-and-Language Navigation (VLN) is an essential skill for embodied agents, allowing them to navigate in 3D environments following natural language instructions. High-performance navigation models require a large amount of training data, the high cost of manually annotating data has seriously hindered this field. Therefore, some previous methods translate trajectory videos into step-by-step instructions for expanding data, but such instructions do not match well with users' communication styles that briefly describe destinations or state specific needs. Moreover, local navigation trajectories overlook global context and high-level task planning. To address these issues, we propose NavRAG, a retrieval-augmented generation (RAG) framework that generates user demand instructions for VLN. NavRAG leverages LLM to build a hierarchical scene description tree for 3D scene understanding from global layout to local details, then simulates various user roles with specific demands to retrieve from the scene tree, generating diverse instructions with LLM. We annotate over 2 million navigation instructions across 861 scenes and evaluate the data quality and navigation performance of trained models. 
The code and dataset are available at 
\href{https://github.com/MrZihan/NavRAG}{https://github.com/MrZihan/NavRAG}

\end{abstract}

\section{Introduction}
Vision-and-Language Navigation (VLN)~\cite{anderson2018vision,krantz2020beyond,qi2020reverie,zhu2021soon} requires the agent to understand natural language instructions and navigate to the described destination in 3D environments. The immense semantic space and diverse forms of language instructions require massive data to train a VLN agent capable of generalizing across different scenarios. However, the high cost of manual annotation has seriously hindered this field, driving efforts to develop instruction generators for automating data generation.

\begin{figure}
\noindent\begin{minipage}[h]{1\columnwidth}%
\begin{center}
\includegraphics[width=1.\columnwidth]{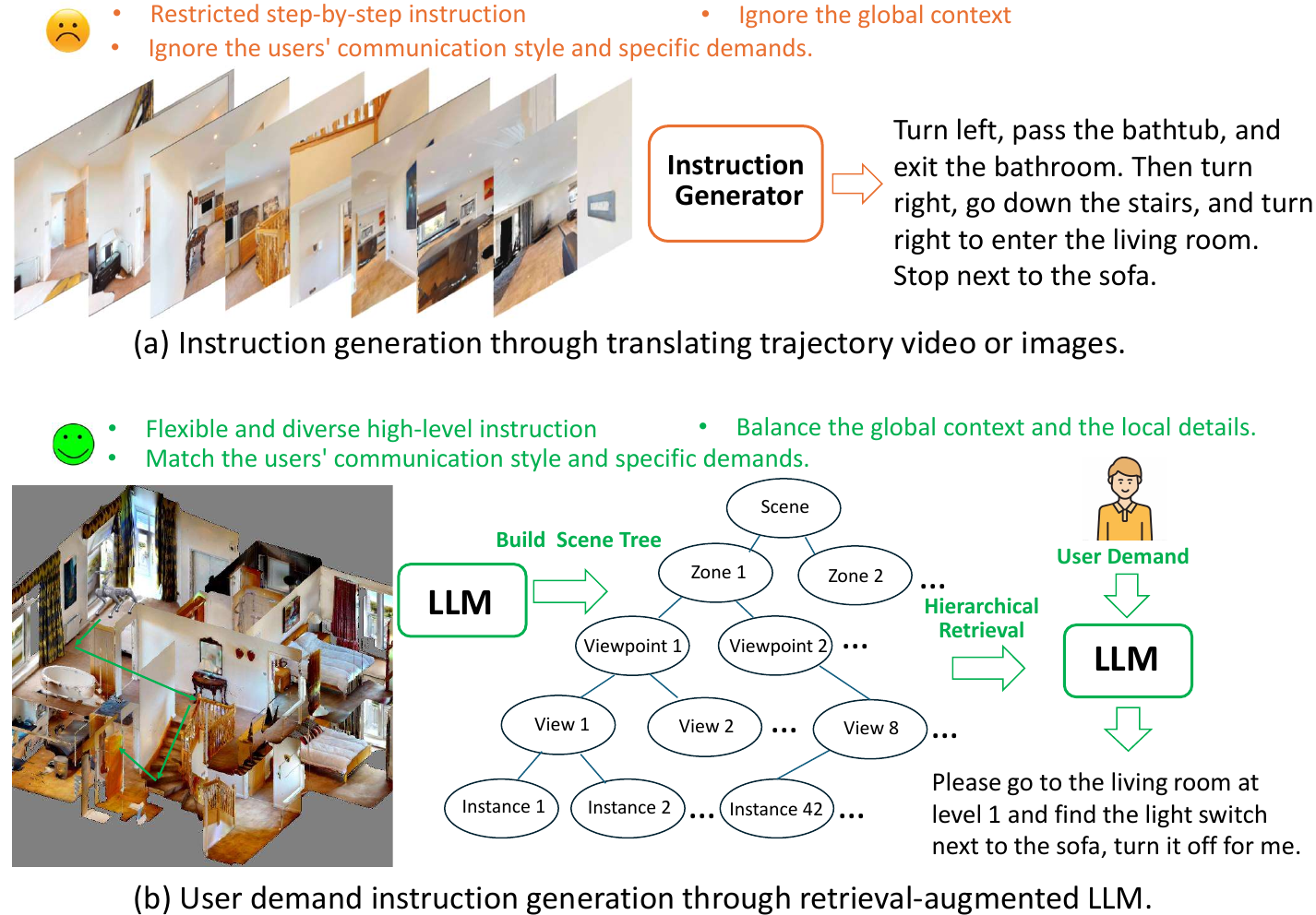}
\par\end{center}%
\end{minipage}
\vspace{-5pt}
\caption{The comparison of previous navigation instruction generation methods (a) and NavRAG (b). 
}
\label{fig:introduction}
\vspace{-6mm}
\end{figure}

As shown in Figure~\ref{fig:introduction} (a), many previous approaches train a navigation instruction generator that takes video or images from Web or simulators as input and produces step-by-step instructions. Leveraging large-scale generated navigation data, this strategy has delivered outstanding results in some navigation tasks using trajectory-based instructions, such as R2R~\cite{anderson2018vision} and REVERIE~\cite{qi2020reverie}. However, such instruction generators still remain some shortcomings:
1) These instruction generators are trained on small-scale and domain-specific datasets, the generated instructions lack diversity; 
2) Such step-by-step instructions are limited to local navigation trajectories overlooking the global context and high-level task planning; 
3) These instructions don't match well with users' natural expressions that describe destinations or state specific needs.

To tackle these challenges, this work proposes NavRAG, an instruction generation method leveraging retrieval-augmented LLM, as illustrated in Figure~\ref{fig:introduction}(b). Specifically, for each 3D scene, NavRAG constructs a scene description tree in a bottom-up manner for hierarchical scene representations. This hierarchical tree comprises multiple layers of language descriptions: the instance layer captures descriptions, attributes, and functionalities of individual instances; the view layer summarizes spatial relationships within a view; the viewpoint layer integrates multiple views into a panoramic environmental description; the zone layer clusters viewpoints within the same functional area (\textit{e.g.}, a bedroom or kitchen); and finally, the scene-level description provides an overview of all zones and their connectivity.

After establishing the environmental context with the scene tree, the generated navigation instructions are expected to \textit{meet the user demands}. Therefore, unlike previous instruction generators that were only used to describe navigation trajectories, NavRAG set up several different user roles (with varying ages, genders, occupations, lifestyles and demands to navigation agent) to simulate and record the instructions sent to navigation agent during one day of this role. Meanwhile, to balance generation quality and cost, our framework initially generates the \textit{coarse} instruction only through the overview of the scene, then uses retrieval-augmented LLM to perform top-down, layer-by-layer retrieval of the best destination and relevant texts from the scene tree, and finally refines the coarse instruction into a more detailed and accurate \textit{refined} instruction using retrieval-augmented LLM.

 In summary, our contributions are as follows:
\begin{itemize}
\item This work proposes an approach for automatically constructing scene description trees and generating user demand navigation instructions using retrieval-augmented LLM.

\item We annotate over 2 million high-quality navigation instructions across 861 3D scenes for training and evaluation.

\item The VLN models trained on our NavRAG dataset achieve superior performance on VLN benchmarks, validating the effectiveness of the proposed method.

\end{itemize}

\section{Related Work}
\noindent \textbf{Vision-and-Language Navigation (VLN)}~\cite{anderson2018vision,krantz2020beyond,qi2020reverie,zhu2021soon} enables embodied agents to navigate to the destination described by the language instructions.
Early VLN researches focus on discrete environments within 90 scenes of Matterport3D~\cite{chang2017matterport3d}, which uses a predefined navigation graph, the agent observes panoramic RGB and depth images, teleporting between graph nodes to follow natural language instructions. Under this setting, the datasets include the step-by-step instruction dataset R2R~\cite{anderson2018vision},  the multilingual instruction dataset RxR~\cite{ku2020room} with longer trajectories, the Remote Embodied Visual Referring Expression (REVERIE)~\cite{qi2020reverie} dataset, and the Scenario Oriented Object Navigation (SOON)~\cite{zhu2021soon} task.
Although efficient for training in discrete environments, these datasets lack real-world applicability. To address this, R2R-CE~\cite{krantz2020beyond} introduce continuous environments~\cite{savva2019habitat} with instructions from the R2R dataset, where agents navigate freely in 3D spaces using low-level actions (e.g., turn 15°, move 0.25m) in the Habitat simulator~\cite{savva2019habitat}. 
In this work, we focus on generating large-scale, high-quality navigation instructions, for simplicity and efficiency, our NavRAG is currently validated in the discrete environments, while the annotated data remains easily transferable to continuous settings.

\noindent \textbf{Navigation Instruction Generation} is an effective approach to addressing the scarcity of training data for VLN. Speaker-follower~\cite{fried2018speaker} and Env-Drop~\cite{tan2019learning} use the LSTM-based instruction generator to generate the offline augmented instructions. VLN-Trans~\cite{zhang2023vln} propose a translator module that enables the navigation agent to generate more concise sub-instructions, leveraging recognizable and distinctive landmarks. 
AutoVLN~\cite{chen2022learning}, MARVAL~\cite{kamath2023new} and ScaleVLN~\cite{wang2023scaling} leverage multiple foundation models~\cite{cheng2022masked,radford2019language,zhaolarge,koh2023simple} and use more 3D scenes to annotate instructions, such as HM3D~\cite{ramakrishnan2habitat} and Gibson~\cite{xia2018gibson}. Recently, more works focus on designing more powerful instruction generator, such as a joint structure for instruction following and generation~\cite{wang2023lana}, Knowledge enhanced speaker~\cite{zeng2023kefa}, LLM instruction generator with chain of thought prompting~\cite{kong2025controllable}, and LLM instruction generator with BEV perception~\cite{fan2025navigation}. However, these methods are limited to identifying landmarks in navigation trajectories and generating low-level instructions, making it difficult to integrate global context, match user demands, and plan high-level tasks. NavRAG will generate navigation instructions better tailored to the application scenario by considering the global context and user demands through scene description trees and retrieval-augmented LLM.

\noindent \textbf{Retrieval-Augmented Generation (RAG)}~\cite{lewis2020retrieval}  was initially introduced to enhance LLMs by retrieving relevant document chunks, thereby providing domain-specific knowledge for better answer. Over time, several innovations have expanded on this idea, including techniques like iterative knowledge retrieval~\cite{shao2023enhancing}, and the incorporation of knowledge graphs~\cite{edge2024local}. Furthermore, adapting RAG to the field of robotics, some works~\cite{xie2024embodied,booker2024embodiedrag} attempt constructing non-parametric memory or scene graphs for 3D scenes, and utilize retrieval-augmented LLM for question answering or navigation. However, traditional RAG methods for scene graph retrieval struggle to balance global context with local details and interpret the environment layout. NavRAG leverages the scene description tree and hierarchical retrieval strategy, achieve better scene understanding.

\section{Method}
\subsection{Navigation Setups}\label{sec:navigation_setups}

\begin{figure*}[ht]
\makebox[\textwidth][c]
{\includegraphics[width=0.7\paperwidth]{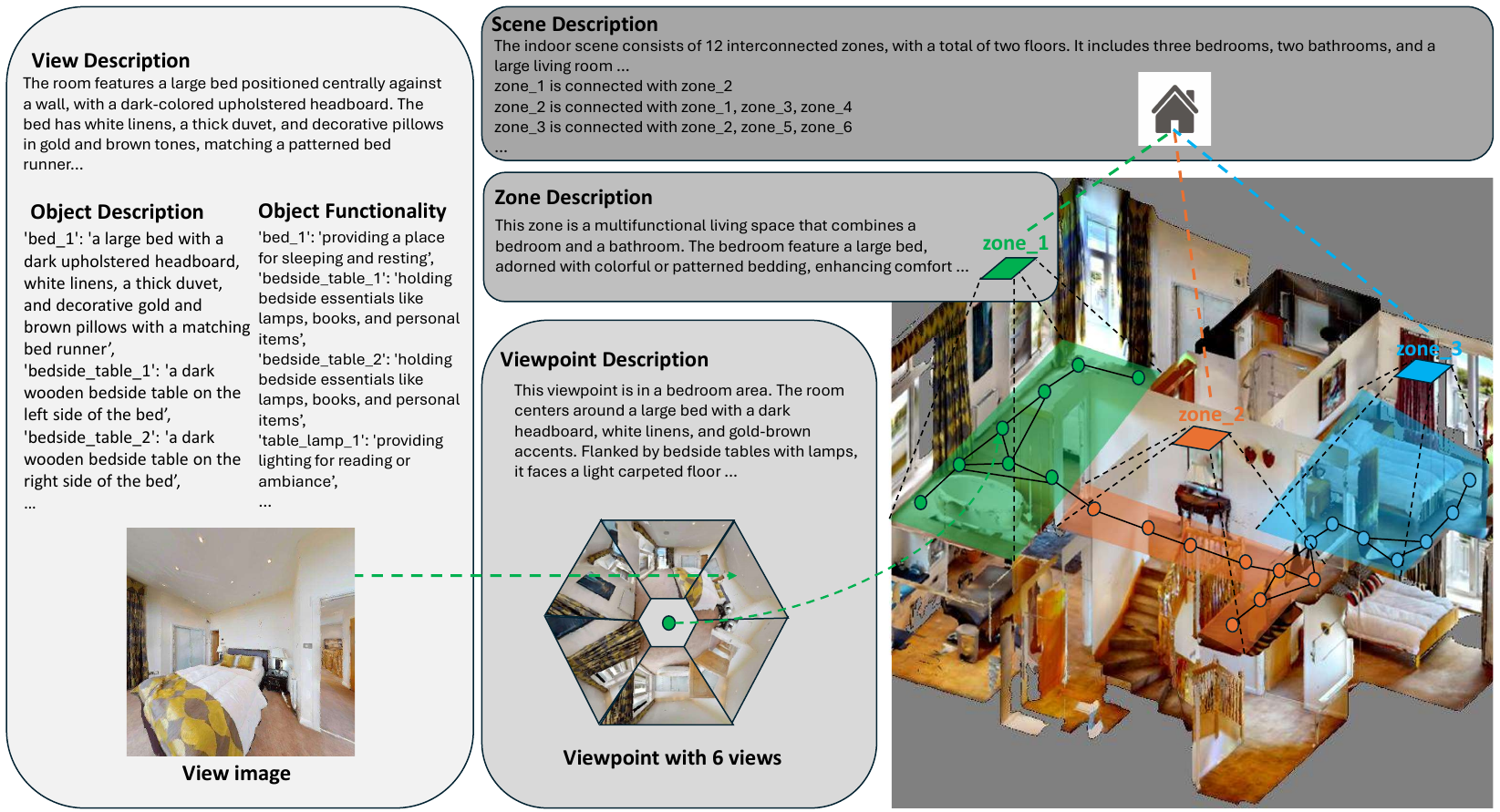}}
\vspace{-20pt}
\caption{Demonstration of the Scene Description Tree. Based on LLM, NavRAG constructs the scene description tree in a bottom-up manner, progressively constructing from objects to views, viewpoints, zones, and the overall scene. This hierarchical structure describes environmental semantics and spatial relationships at different levels, facilitating LLM in understanding 3D environments and retrieving information for instruction generation.}
\label{fig:framework}
\vspace{-10pt}
\end{figure*}

In the vision-and-language navigation (VLN) setting, the navigation connectivity graph \(\mathcal{G} = \{\mathcal{V}, \mathcal{E}\}\) is provided by the Matterport3D simulator~\cite{chang2017matterport3d}, where \(\mathcal{V}\) represents navigable nodes and \(\mathcal{E}\) denotes the edges connecting them. The agent is equipped with RGB cameras and a GPS sensor. Starting from a starting node and following natural language instructions, the agent must explore the navigation connectivity graph \(\mathcal{G}\) and move to the destination node. The instruction is represented by a sequence of word embeddings \(\mathcal{W} = \{w_l\}_{l=1}^{L}\), where \(L\) is the number of words. At each time step \(t\), the agent can perceive a panoramic RGB observation \(\mathcal{R}_{t} = \{r_{t,k}\}_{k=1}^{K}\) at current node \(\mathcal{V}_{t}\), consisting of \(K\) view images. The RGB observation of nodes can be obtained through the Matterport3D simulator, so each annotated navigation sample only needs a navigation instruction and an optimal path from the starting node to the destination node for training or evaluation.

\subsection{Constructing the Scene Description Tree}\label{sec:scene_tree}
Before generating instructions, it is essential to first represent and understand the environment. As illustrated in Figure~\ref{fig:framework}, we propose a bottom-up, hierarchical approach for constructing a scene description tree. At the view and object levels, each object is described with fine-grained details, including its category, attributes, functionality. The spatial relations among objects is summarized in view-level description. The viewpoint level aggregates multiple views surrounding each navigable viewpoint and summarize the spatial layout around this viewpoint. The zone level integrates multiple viewpoints to define large functional areas (\textit{e.g.}, a bedroom) within the 3D scene. Finally, the house level encompasses multiple zones, offering a high-level abstraction of the overall spatial layout and functional partitioning of the whole scene.

\noindent \textbf{Navigation Graph.} We introduce 800 training scenes from HM3D~\cite{ramakrishnan2habitat} and 61 training scenes along with 11 validation scenes from Matterport3D~\cite{chang2017matterport3d} for scene tree construction. Obtaining the navigation graphs of these scenes is the first step. Although MP3D already has manually annotated navigation graphs, the navigation graphs of HM3D still remains to construct. Following ScaleVLN~\cite{wang2023scaling}, we use a heuristic method to build high-quality navigation graphs for HM3D scenes, ensuring high space coverage, fully traversable edges, and well-positioned nodes, which samples dense viewpoints using Habitat Simulator~\cite{savva2019habitat}’s navigable position function, ensuring over 0.4m geodesic separation. The Agglomerative Clustering (1.0m threshold) is utilized to centralize nodes and form an initial graph by randomly connecting viewpoints within 5.0m, capping node edges at five. Finally, the graph is refined for full connectivity and traversal, producing graphs for 800 scenes.

\noindent \textbf{View and Object-level Annotation.} To capture detailed information about objects within a specific viewpoint of the navigation graph, we utilize the Habitat simulator~\cite{savva2019habitat} to uniformly sample six views (each with an image resolution of 480×480) from every viewpoint in the navigation graph. These views are then input into a multimodal LLM (\textit{i.e.}, GPT-4o-mini~\cite{hurst2024gpt}) to generate descriptions of each view, objects, their attributes, and functionalities.

\noindent \textbf{Viewpoint-level Annotation.} 
Integrating descriptions and object information from multiple views, the LLM generates a comprehensive description of the environment surrounding the viewpoint. This description encompasses the area type, spatial layout, and relationships among objects, providing a holistic understanding of panorama.

\begin{figure}
\noindent\begin{minipage}[h]{1\columnwidth}%
\begin{center}
\includegraphics[width=1\columnwidth]{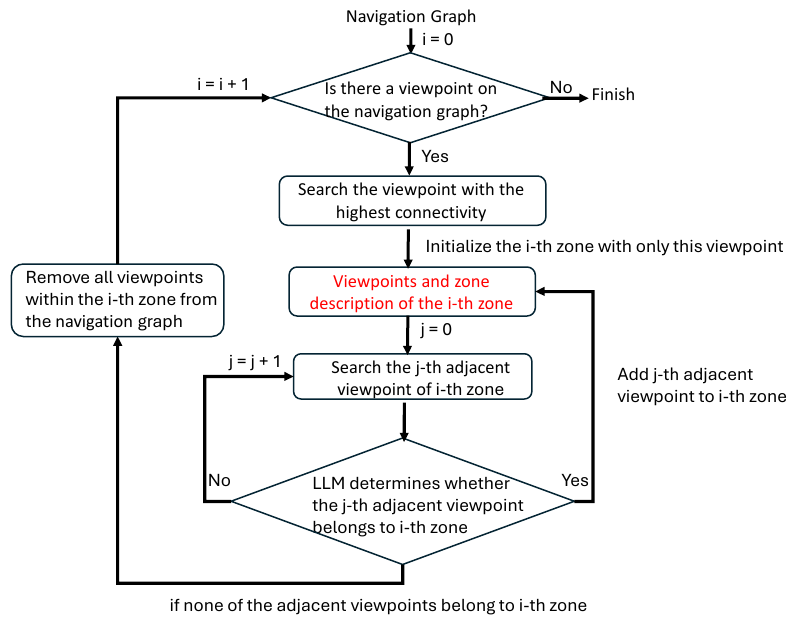}
\par\end{center}%
\end{minipage}
\vspace{-5pt}
\caption{Framework of the zone partitioning algorithm based on connectivity relations and environmental semantics. 
}
\label{fig:zone_level}
\vspace{-6mm}
\end{figure}

\noindent \textbf{Zone Partitioning and Annotation.} 
To enhance the comprehension of the scene's spatial layout (e.g., room count and connectivity) meanwhile decreasing retrieval cost from numerous viewpoints, we construct zones by merging multiple viewpoints, as shown in Figure~\ref{fig:framework}. Unlike previous methods~\cite{xie2024embodied} using hierarchical clustering based on spatial positions to construct scene trees, we propose a new algorithm that incorporates viewpoint connectivity and environmental semantics for scene partitioning as shown in Figure~\ref{fig:zone_level}. Hierarchical clustering based on spatial positions has two important drawbacks:  
1) It overlooks viewpoint connectivity, potentially grouping nearby but wall-separated viewpoints into the same zone.  
2) It ignores environmental semantics, relying solely on spatial positions cannot accurately recognize different functional areas of the scene. 

To address these issues, our algorithm first selects the viewpoint with the highest connectivity to initialize a zone and uses LLM to generate its description. Then, by searching the adjacent viewpoints in descending order of connectivity, the algorithm inputs the zone description and the description of adjacent viewpoint into LLM to determine if the viewpoint belongs to the zone, if yes, this viewpoint will be added to the zone, and the zone description is updated. Once all viewpoints for this zone are identified, all nodes within the zone are removed from the navigation graph, then the next zone construction begins.

\noindent \textbf{Scene-level Annotation.} 
To provide an overview of the spatial layout of the entire scene, the scene-level description primarily includes the connectivity between various zones (similar to MapGPT~\cite{chen2024mapgpt}), the types of each zone, a concise summary, and the functionality.

\begin{figure*}[h]
\makebox[\textwidth][c]
{\includegraphics[width=0.7\paperwidth]{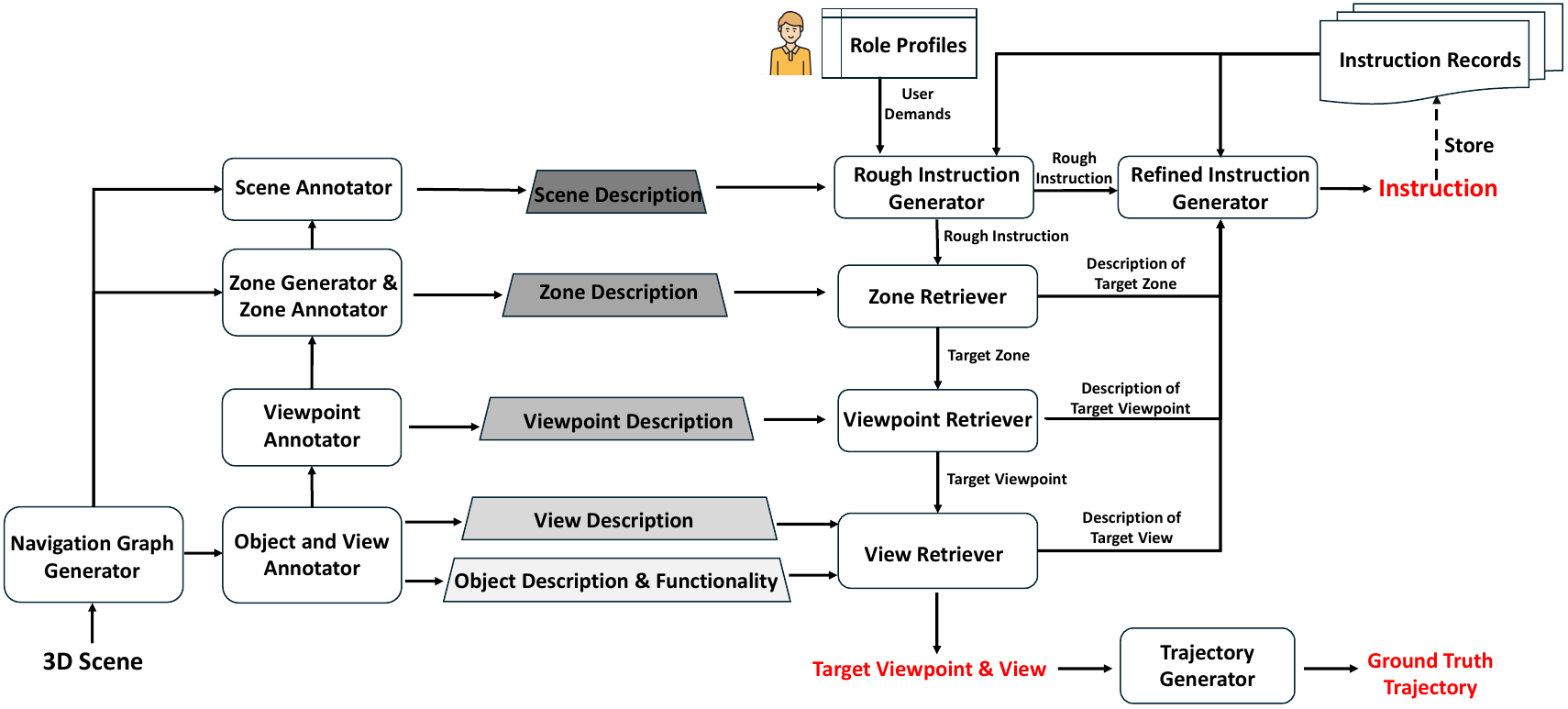}}
\vspace{-22pt}
\caption{Framework of NavRAG for scene tree construction and navigation instruction generation through Retrieval-Augmented LLM.}
\label{fig:framework_2}
\vspace{-10pt}
\end{figure*}

\subsection{User Demand Instruction Generation}
\begin{figure*}[h]
\makebox[\textwidth][c]
{\includegraphics[width=0.65\paperwidth]{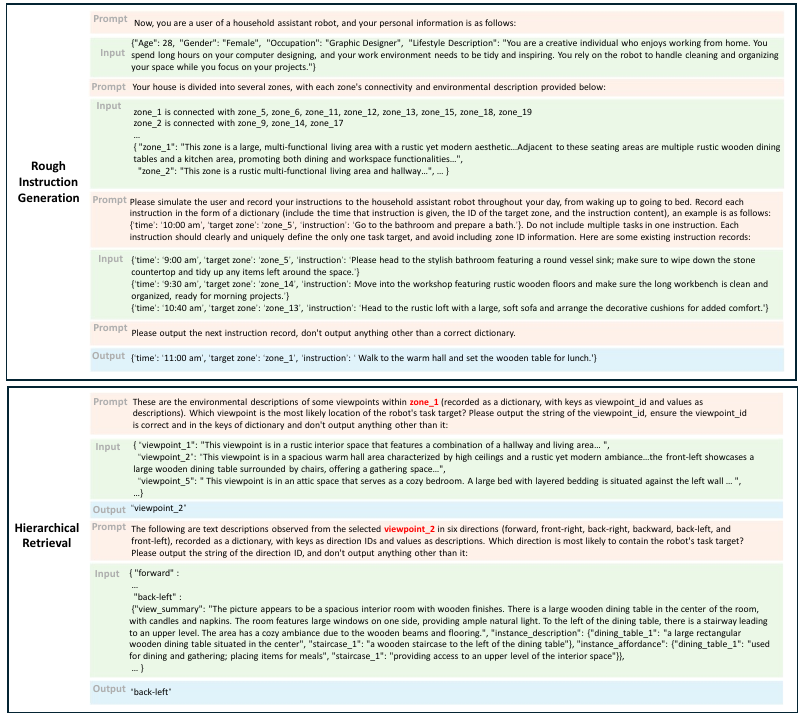}}
\vspace{-20pt}
\caption{Prompt, input and output of the Rough Instruction Generator and Hierarchical Retrieval.}
\label{fig:prompt}
\vspace{-5pt}
\end{figure*}

As shown in Figure~\ref{fig:framework_2} and Figure~\ref{fig:prompt}, after constructing the scene description tree, NavRAG leverages the scene-level description, user information, and demands to generate a rough instruction for the navigation agent, such as "Walk to the warm hall and set the wooden table for lunch". Subsequently, NavRAG performs a top-down, hierarchical retrieval of potential destinations from the scene tree and integrates the retrieved environmental descriptions at different levels into the LLM, to refine rough instruction into precise and comprehensive instruction, such as "Walk to the warm hall featuring elegant wooden accents and set the large wooden table with candles
and napkins for a lovely dinner ambiance". 

\noindent \textbf{User Demands Simulation.} 
To further improve the diversity of generated instructions and meet the user demands, NavRAG integrates texts of user information and demands, enabling the instruction generator to simulate specific roles to generate tailored instructions. A sample of user profile and demands is as follows:{\small{\\
\noindent\{ \\
\indent"Age": 33,  \\
\indent"Gender": "Female", \\
\indent"Occupation": "Lawyer",  \\
\indent"Lifestyle Description": "You maintain the good habit of going to bed early and waking up early. Besides working in the study, you often do yoga and other exercises in the living room and enjoy cooking your own meals." \\
\}
}}

We manually annotate 20 user profiles for different roles. For each role, the prompt guides the LLM in simulating the role's behavior with a given scene description tree, generating the records of 50 navigation instructions sent to the agent during one day of this role.

\begin{figure*}[h]
\makebox[\textwidth][c]
{\includegraphics[width=0.65\paperwidth]{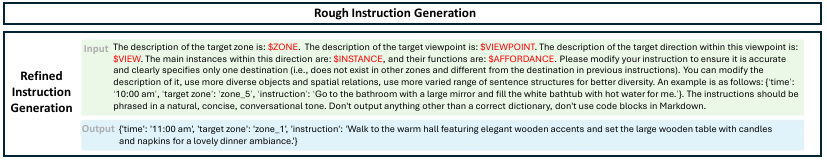}}
\vspace{-20pt}
\caption{Prompt, input and output of the Refined Instruction Generator. \$ZONE, \$VIEWPOINT, \$VIEW, \$INSTANCE and \$AFFORDANCE denote retrieved environmental descriptions at different levels.}
\label{fig:prompt_2}
\vspace{-10pt}
\end{figure*}

\noindent \textbf{Retrieval-Augmented Generation.} As illustrated in Figure~\ref{fig:framework_2}, Figure~\ref{fig:prompt} and Figure~\ref{fig:prompt_2}, NavRAG performs layer-by-layer retrieval of texts at different levels based on the scene description tree, progressively localizes the navigation destination. Initially, the LLM generates a rough instruction based on scene-level descriptions, user information, and historical instruction records. It then identifies the most probable zone containing the navigation destination from the zone-level descriptions. Based on the viewpoint descriptions within that zone, the LLM selects the target viewpoint and locates the view containing the navigation target. By integrating texts from all different levels, the LLM ultimately refines rough instruction and outputs the precise and comprehensive instruction.


\section{Experiments}

\begin{table}[ht]
\vspace{-5pt}
\tabcolsep=0.04cm
\centering{}%
\scriptsize
\begin{tabular}{c|c|ccc}
\hline 
\multirow{1}{*}{Dataset}  & \multirow{1}{*}{Generated}  & \multirow{1}{*}{\#Scene} & \multirow{1}{*}{\#Instr.} & \multirow{1}{*}{Instr. length}
\tabularnewline
\hline 

REVERIE~\cite{qi2020reverie} & $\times$ & 60 & 10,466 & 18.64 
\tabularnewline

R2R~\cite{anderson2018vision} & $\times$ & 61 & 14,039 & 26.33  
\tabularnewline

RxR-en~\cite{ku2020room} & $\times$ & 60 & 26,464 & 102.13  
\tabularnewline

SOON~\cite{zhu2021soon} & $\times$ & 34 &  2,780 & 44.09  
\tabularnewline

Prevalent~\cite{hao2020towards} & \ding{51} & 60 & 1,069,620 & 24.23 
\tabularnewline

Marky~\cite{wang2022less} & \ding{51} & 60 & 333,777 & 99.45 
\tabularnewline

AutoVLN~\cite{chen2022learning} & \ding{51} & 900 & 217,703 & 20.52 
\tabularnewline

ScaleVLN~\cite{wang2023scaling} & \ding{51} & 1289 & 4,941,710 & 21.61 
\tabularnewline

NavRAG (Ours) & \ding{51} & 861 & 2,115,019 & 29.11 
\tabularnewline

\hline 
\end{tabular}
\vspace{-5pt}
\caption{Statistics of training data on different VLN datasets.}\label{datasets}
\vspace{-5mm}
\end{table}

\subsection{Datasets and Evaluation Metrics}
\noindent \textbf{Datasets.} Table~\ref{datasets} summarizes the main VLN datasets, including human-annotated data and model-generated data. The high cost of manual annotation limits the scale of manual training data, severely restricting the generalization ability of VLN models. An effective approach to enhancing navigation performance is to automatically generate large-scale navigation data for VLN pretraining, then fine-tune on manual data. 
Our NavRAG annotates over 2 million navigation instructions across 861 training scenes, each corresponding to a navigation destination (\textit{i.e.}, target viewpoint). Using a trajectory generator which samples the starting viewpoint and calculate the shortest path to the destination, we randomly sample 5 trajectories per instruction, yielding over 10 million navigation trajectories in total. To evaluate model performance, we also annotate 7,396 instruction-trajectory pairs across 11 unseen scenes, forming the \textit{NavRAG Val Unseen} benchmark for performance evaluation.

\noindent \textbf{Evaluation Metrics.} Four main metrics are used for navigation: 1) Navigation
 Error (NE): the mean of the shortest path distance between the agent’s final position and the destination.
 2) Oracle Success Rate (OSR): the percentage that the agent has reached a position within 3 meters of the destination. 3) Success Rate (SR): the percentage of the predicted stop position being within 3 meters from the destination. (3) Success rate weighted Path Length (SPL) that normalizes the success rate with trajectory length.

\begin{table*}[h]
\small
\tabcolsep=0.07cm
\centering{}%
\begin{tabular}{c|c|c|cccc|cccc}
\hline 
\multirow{2}{*}{Models} & \multirow{2}{*}{LLM} & \multirow{2}{*}{Training Data} & \multicolumn{4}{c|}{NavRAG Val Unseen} & \multicolumn{4}{c}{REVERIE Val Unseen}\tabularnewline
\cline{4-11} \cline{5-11} \cline{6-11} \cline{7-11} \cline{8-11} \cline{9-11} \cline{10-11} \cline{11-11}
 & & & NE\textdownarrow{} & OSR\textuparrow{} & SR\textuparrow{} & SPL\textuparrow{} & NE\textdownarrow{} & OSR\textuparrow{} & SR\textuparrow{} & SPL\textuparrow{}\tabularnewline
\hline 
\textcolor{black} DUET & $\times$ & AutoVLN (REVERIE-style) & 13.2 & 30.2 & 16.2 & 10.7 & \textcolor{gray}{6.9} & \textcolor{gray}{49.7} & \textcolor{gray}{42.3} & \textcolor{gray}{26.4} 
 \tabularnewline

  \textcolor{black} DUET & $\times$ & ScaleVLN (REVERIE-style) & 11.3 & 41.9 & 17.4 & 11.9 & \textcolor{gray}{6.7} & \textcolor{gray}{50.2} & \textcolor{gray}{44.6} & \textcolor{gray}{28.2} 
 \tabularnewline

  \textcolor{black} DUET & $\times$ & ScaleVLN (R2R-style) & 12.6 & 31.1 & 10.3 & 4.2 & 9.0 & 41.4 & 27.9 & 11.6 
 \tabularnewline
 
\textcolor{black} NavGPT & GPT-4o-mini & - & \textbf{7.7} & 43.1 & 28.2 & 11.6 & 9.2 & 25.8 & 20.2 & 13.1 
 \tabularnewline

\textcolor{black} MapGPT & GPT-4o-mini & - & 7.8 & \textbf{47.7} & \textbf{30.9} & \textbf{15.3} & 8.2 & 37.4 & 30.2 & 21.6 
 \tabularnewline

\textcolor{black} MapGPT & Llama-3.1-8B~\cite{dubey2024llama} & - & 8.1 & 44.2 & 25.5 & 12.4 & 8.4 & 35.8 & 24.4 & 16.2 
 \tabularnewline
 
\hline 

\textcolor{black} HAMT & $\times$ & NavRAG (Ours) & \textcolor{gray}{8.3} & \textcolor{gray}{42.5} & \textcolor{gray}{25.1} & \textcolor{gray}{20.4} & 8.1 & 40.3 & 32.8 & 21.7 
 \tabularnewline
 
 \textcolor{black} DUET & $\times$ & NavRAG (Ours) & \textcolor{gray}{7.7} & \textcolor{gray}{50.0} & \textcolor{gray}{30.7} & \textcolor{gray}{25.4} & \textbf{7.6} & \textbf{45.9} & \textbf{36.1} & \textbf{24.9}
 \tabularnewline

 \hline
\end{tabular}
\vspace{-8pt}
\caption{Zero-shot performance comparison on NavRAG and REVERIE datasets, reflecting the model's generalization ability. Gray values do not strictly follow the zero-shot setting.}\label{zero-shot}
\vspace{-5pt}
\end{table*}

\subsection{VLN Models}
To evaluate our NavRAG dataset, multiple VLN models are used in the experiments, as shown in Table~\ref{zero-shot} and Table~\ref{sota_compare}.

\noindent \textbf{DUET} (Dual-scale Graph Transformer)~\cite{chen2022think} is a VLN model that dynamically builds a topological map for efficient global exploration while integrating fine-grained local observations and coarse-grained global encoding through graph transformers. 

\noindent \textbf{HAMT} (History Aware Multimodal Transformer)~\cite{chen2021history} is a VLN model that integrates long-horizon history using a hierarchical vision transformer, which efficiently encodes past panoramic observations and combines text, history, and current views to predict navigation actions.

\noindent \textbf{NavGPT}~\cite{zhou2024navgpt} is a purely LLM-based instruction-following navigation agent, which performs zero-shot sequential action prediction, demonstrating abilities such as high-level planning, sub-goal decomposition, commonsense integration, and navigation progress tracking.

\noindent \textbf{MapGPT}~\cite{chen2024mapgpt} is a LLM-based VLN agent that integrates an online linguistic-formed map to enable global exploration. By incorporating node information and topological relationships into prompts, MapGPT understands spatial environments and features an adaptive planning mechanism for multi-step path planning.

\subsection{Limitations of the Existing Training Data.}
Table~\ref{zero-shot} evaluates the zero-shot performance of multiple VLN methods on NavRAG and REVERIE benchmarks, and also shows the performance of models trained on NavRAG datasets. As shown in rows 1-3 of Table~\ref{zero-shot}, models trained on previously generated large-scale datasets (\textit{i.e.}, AutoVLN and ScaleVLN) perform poorly on the NavRAG benchmark, whereas LLM-based methods (rows 4-6) demonstrate relatively strong performance. 

NavRAG leverages the scene description tree and retrieval-augmented LLM, resulting in a larger semantic space of instructions with more diverse sentence structures, meanwhile, better aligned with human expression. LLM-based models effectively comprehend these instructions. In contrast, instructions in ScaleVLN and AutoVLN are generated by a pre-trained instruction generator trained on a small-scale manually annotated dataset (\textit{i.e.} REVERIE and R2R), restricting the semantic space and diversity, and further hindering the generation ability. Thus, models trained on them struggle with NavRAG benchmark and real-world applications.

Notably, the performance of the LLM-based method on the NavRAG benchmark surpasses the human-annotated REVERIE benchmark (NE, OSR and SR metrics), due to NavRAG's longer, more detailed, and accurate instructions (shown in  Table~\ref{datasets}). This finding further validates the quality of instructions generated by our NavRAG.

\subsection{Generalization Ability of NavRAG}

As shown in the last two rows of Table~\ref{zero-shot}, the models trained on the NavRAG dataset achieves competitive performance on both NavRAG Val Unseen and REVERIE Val Unseen benchmarks, and even outperforms LLM-based methods (\textit{i.e.}, NavGPT and MapGPT), showing the ability of NavRAG dataset to enhance model generalization.

\begin{figure}
\noindent\begin{minipage}[h]{1\columnwidth}%
\begin{center}
\includegraphics[width=1\columnwidth]{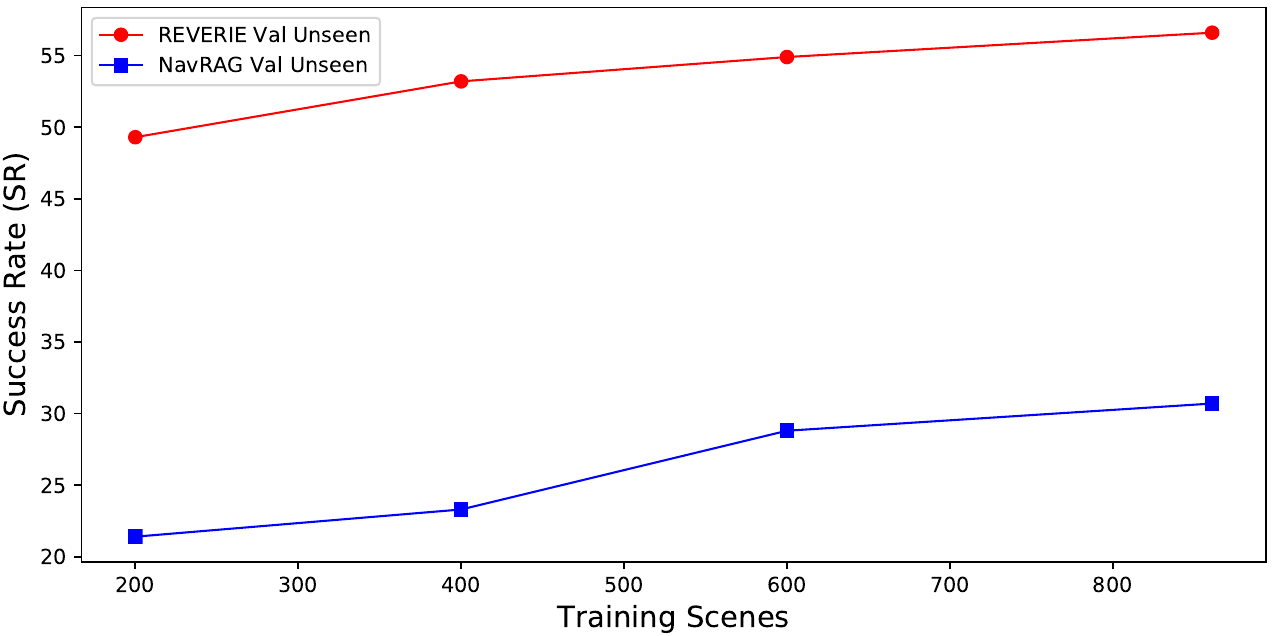}
\par\end{center}%
\end{minipage}
\vspace{-8pt}
\caption{Navigation performance with respect
 to the number of pre-training scenes in
 NavRAG dataset.}
\label{fig:scene_count}
\vspace{-6mm}
\end{figure}

Furthermore, Figure~\ref{fig:scene_count} illustrates that NavRAG consistently improves the performance of the VLN model as the pre-training data scale increases, underscoring the potential and value of large-scale generated navigation data.

\begin{table}[ht]
\vspace{-5pt}
\scriptsize
\tabcolsep=0.04cm
\centering{}%
\begin{tabular}{c|c|c|cccc}
\hline 
\multirow{2}{*}{Methods} & \multirow{2}{*}{LLM} & \multirow{2}{*}{Objects}  &\multicolumn{4}{c}{REVERIE Val Unseen}\tabularnewline
\cline{4-7} \cline{5-7} \cline{6-7} \cline{7-7} 
  & & & NE\textdownarrow{} & OSR\textuparrow{} & SR\textuparrow{} & SPL\textuparrow{} \tabularnewline
\hline

HAMT~\cite{chen2021history} & $\times$ & \ding{51} & - & 36.8 & 33.0 & 30.2
\tabularnewline

DUET~\cite{chen2022think} & $\times$ & \ding{51} & - & 51.1 & 47.0 & 33.7
\tabularnewline

Lily~\cite{lin2023learning} & $\times$ & \ding{51} & - & 53.7 & 48.1 & 34.4
\tabularnewline

KERM~\cite{li2023kerm} & $\times$ & \ding{51} & - & 55.2 & 50.4 & 35.4
\tabularnewline

BEVBert~\cite{an2023bevbert} & $\times$ & \ding{51} & - & 56.4 & 51.8 & 36.4
\tabularnewline

BSG~\cite{liu2023bird} & $\times$ & \ding{51} & - & 58.1 & 52.1 & 35.6
\tabularnewline

GridMM~\cite{wang2023gridmm} & $\times$ & \ding{51} & - & 57.5 & 51.4 & 36.5
\tabularnewline

ENP-DUET~\cite{liu2024vision} & $\times$ & \ding{51} & - &  54.7 & 48.9 & 33.8
\tabularnewline

AutoVLN~\cite{chen2022learning} & $\times$ & \ding{51} & - & 62.1 & 55.9 & 40.9
\tabularnewline

ScaleVLN~\cite{wang2023scaling} & $\times$ & \ding{51} & - & \textbf{63.9} & 57.0 & 41.8
\tabularnewline

VER~\cite{liu2024volumetric} & $\times$ & \ding{51} & - & 61.1 & 56.0 & 39.7
\tabularnewline

GOAT~\cite{wang2024vision} & $\times$ & \ding{51} & - & - & 53.4 & 36.7
\tabularnewline

NaviLLM~\cite{zheng2024towards} & \ding{51} & \ding{51} & - & 53.7 & 44.6 & 36.6
\tabularnewline

MiC~\cite{qiao2023march} & \ding{51} & \ding{51} & - & 62.4 & 57.0 & 43.6
\tabularnewline

VLN-Copilot~\cite{qiao2024llm} & \ding{51} & \ding{51} & - & 62.6 & \textbf{57.4} & \textbf{43.6}
\tabularnewline

\hline
DUET & $\times$ & $\times$ & 6.0 & 50.0 & 45.8 & 32.5
\tabularnewline

AutoVLN & $\times$ & $\times$ &5.7 & 61.8 & 54.3 & 39.1
\tabularnewline

ScaleVLN & $\times$ & $\times$ &5.7 & 62.7 & 55.9 & 40.6
\tabularnewline

NavRAG (Ours) & $\times$ & $\times$ & \textbf{5.5} & \textbf{70.7} & \textbf{57.3} & \textbf{42.0}
\tabularnewline

\hline 
\end{tabular}
\vspace{-5pt}
\caption{Fine-tuning performance comparison on REVERIE dataset. "Objects" indicates whether visual features of annotated object bounding boxes are utilized for training.}\label{sota_compare}
\vspace{-4mm}
\end{table}

\subsection{Comparison with SOTA Methods}

The last row of Table~\ref{sota_compare} presents the performance of DUET pre-trained on the NavRAG dataset and fine-tuned on the REVERIE dataset, which is comparable to the SOTA approaches with LLM.

Previous methods use manually annotated object bounding boxes of REVERIE datasets to extract visual features for model inputs. However, this strategy restricts the model's applicability in real-world deployment, since the real world does not have ground-truth object information. NavRAG removes the reliance on annotated object bounding boxes, making it more suitable for real-world deployment. For a fair comparison, we also evaluate the performance of other generated datasets after removing the object bounding box information from REVERIE, in this setting, NavRAG shows superior performance. This suggests that, despite NavRAG having a larger domain gap with the REVERIE dataset compared to AutoVLN and ScaleVLN, pretraining on more diverse instructions of the NavRAG dataset enables the model to achieve strong generalization, even leading to better fine-tuning performance surpasses domain-specific generated data.

\subsection{Ablation Study}

\begin{table}[ht]
\vspace{-5pt}
\small
\tabcolsep=0.04cm
\centering{}%
\begin{tabular}{c|c|cccc}
\hline 
\multirow{2}{*}{Training Data}  & \multirow{2}{*}{Validation Data}  &\multicolumn{4}{c}{NavRAG Val Unseen}\tabularnewline
\cline{3-6} \cline{4-6} \cline{5-6} \cline{6-6} 
  & & NE\textdownarrow{} & OSR\textuparrow{} & SR\textuparrow{} & SPL\textuparrow{} \tabularnewline
\hline 

GraphRAG & GraphRAG & 14.1 & 41.4 & 12.1 & 8.7
\tabularnewline

\hline

Zone Clustering & Zone Clustering & 9.8 & 48.9 & 16.4 & 11.6 
\tabularnewline

\hline

w/o User. & w/ User. & 9.4 & 45.6 & 18.6 & 13.7 
\tabularnewline

w/ User. & w/o User. & 9.1 & 48.1 & 20.8 & 15.7
\tabularnewline

\hline

NavRAG & NavRAG & 8.9 & 46.8 & 21.5 & 15.4
\tabularnewline

\hline 
\end{tabular}
\vspace{-5pt}
\caption{The ablation study of NavRAG, evaluating the effectiveness of the components. To reduce costs, only 100 scenes are annotated for DUET training.}\label{ablation_modules}
\vspace{-2mm}
\end{table}

\noindent \textbf{Retrieval-Augmented Generation: NavRAG vs. GraphRAG.} To validate the superiority of our scene description tree-based retrieval over traditional RAG methods (\textit{e.g.}, GraphRAG~\cite{edge2024local}), we also annotate 100 scenes through GraphRAG to evaluate instruction quality. Specifically, GraphRAG replaces the scene description tree with a knowledge graph built from view-level descriptions. During instruction generation, it retrieves relevant text fragments from the knowledge graph, integrates them into a prompt, and feeds them to the LLM to generate instructions and navigation destinations. Comparing the first and last rows of Table~\ref{ablation_modules} shows that the model trained with GraphRAG-annotated data performs poorly on its validation set, indicating low annotation quality.

\noindent \textbf{Zone Partitioning Algorithm.} Row 2 of Table~\ref{ablation_modules} evaluates the instruction quality using zones from hierarchical clustering~\cite{xie2024embodied}. Compared to our proposed zone partitioning algorithm, hierarchical clustering relies solely on the distance between different viewpoints, disregarding the spatial layout of the environment (\textit{e.g.}, wall partitions) and lacking environmental semantic understanding.

\noindent \textbf{Role Simulation and User Demands.} 
To enhance the diversity of instructions and better match user demands, we design prompts that guide the LLM to simulate a user with a specific role profile and generate instructions to the agent in everyday scenarios. 
As shown in rows 3 and 4 of Table 5, we analyze the impact of role simulation and user demands on the quality of NavRAG-generated instructions. When user demands are not utilized for training data generation, performance significantly decreases in validation data with diverse user demands (Table 5, row 3). However, if user demands are included in the training data but removed from the validation data, the model still maintains strong performance. The experimental results indicate that enhancing the diversity of generated instructions by simulating user roles and incorporating user demands is feasible. Moreover, more diverse instructions can provide the model with stronger generalizability and performance.

\section{Conclusion}

In this work, we propose NavRAG, a user demand-oriented navigation data generation method through retrieval-augmented LLM. Unlike previous works that use trajectory-based instruction generators to translate navigation videos into step-by-step instructions, our NavRAG utilizes the environmental representations from a hierarchical scene description tree. By retrieving descriptions of different levels in a top-down manner and introducing the user demands, NavRAG effectively enhances the quality of instructions generated by the LLM. 

\section{Limitations}
1) Although the strong navigation performance shows the quality of the NavRAG dataset, no effective method exists to evaluate the correctness of generated instructions. Previous approaches evaluate instruction generators by comparing generated instructions with human-annotated instructions (\textit{e.g.}, using metrics like Bleu, SPICE, and CIDEr). However, our experiments show that small-scale human annotations lack diversity and are insufficient for accurately evaluating dataset quality. 2) The navigation targets annotated by NavRAG are limited to the viewpoint-level, failing to precisely locate specific target objects and their positions, which restricts its applicability in object-centered tasks such as mobile manipulation.

\bibliography{custom}

\end{document}